\documentclass[11pt,a4paper]{article}
\usepackage[hyperref]{emnlp2020}
\usepackage{times}
\usepackage{latexsym}
\usepackage{microtype}

\usepackage{xcolor}
\usepackage{graphicx}
\usepackage{makecell}
\usepackage{url}
\usepackage{xr}
\usepackage{amsmath, amsthm, amssymb}
\usepackage{titlecaps}
\usepackage{caption}
\usepackage{natbib}
\renewenvironment{thebibliography}[1]{
  \begin{oldthebibliography}{#1}
    \setlength{\itemsep}{1em}
    \setlength{\parskip}{0em}
}
{
  \end{oldthebibliography}
}

\usepackage{subfig}
\makeatletter
\newcommand*{\addFileDependency}[1]{
  \typeout{(#1)}
  \@addtofilelist{#1}
  \IfFileExists{#1}{}{\typeout{No file #1.}}
}
\makeatother
 
\newcommand*{\myexternaldocument}[1]{%
    \externaldocument{#1}%
    \addFileDependency{#1.tex}%
    \addFileDependency{#1.aux}%
}
\myexternaldocument{appendix}

\usepackage{microtype}

\aclfinalcopy 

\definecolor{darkgreen}{rgb}{0,0.6,0}
\definecolor{darkblue}{rgb}{0,0,0.6}

\newcommand{\webred}{$\mathtt{\mathbf{W}ebRED}{\sc }$}
\newcommand{\htwentyone}{$\mathtt{\mathbf{W}ebRED}_{H\texttt{2+1}}$}
\newcommand{\hfive}{$\mathtt{\mathbf{W}ebRED}_{H\texttt{5}}$}
\newcommand{\ds}{$\mathtt{\mathbf{W}ebRED}_{DS}$}

\title{\titlecap{WebRED: Effective pretraining and finetuning for relation extraction on the web}}

\author{Robert Ormandi \\
  Google Brain \\ \And
  Mohammad Saleh \\
  Google Brain \\
  {\hspace{10em} \tt \{ormandi*,msaleh,erinmw,vinaysrao\}@google.com} \\
  \And
  Erin Winter \\
  Google \\ \And
  Vinay Rao \\
  Google Brain \\
}

\date{}

\begin{document}
\maketitle
\begin{abstract}
Relation extraction is used to populate knowledge bases that are important to many applications. Prior datasets used to train relation extraction models either suffer from noisy labels due to distant supervision, are limited to certain domains or are too small to train high-capacity models. This constrains downstream applications of relation extraction. 
We therefore introduce: \webred{} (Web Relation Extraction Dataset), a strongly-supervised human annotated dataset for extracting relationships from a variety of text found on the World Wide Web, consisting of $\sim$110K examples. We also describe the methods we used to collect $\sim$200M examples as pre-training data for this task.
We show that combining pre-training on a large weakly supervised dataset with fine-tuning on a small strongly-supervised dataset leads to better relation extraction performance. We provide baselines for this new dataset and present a case for the importance of human annotation in improving the performance of relation extraction from text found on the web. 
\end{abstract}

\section{Introduction} \label{sec:motivation}

Relationship extraction is the task of extracting semantic relationships from a text. Such a relationship occurs between one or more entities of a certain type (eg: person, organization) and belongs to a particular semantic category (eg: date of birth, employed by). Consider the sentence ``Alice lives in Baltimore''. Here, the relation `lives in' connects the subject entity `Alice' to the object entity `Baltimore'. Relation extraction has many applications in information extraction, creating or extending knowledge bases, automatically annotating structured information found in text and recently, in evaluating the factual consistency of abstractive text summarization \citep{factualaccuracy,evaluating_factual_consistency,optimizing_factual_correctness}.

Typically, datasets for relationship extraction are constructed using distant supervision \citep{distantsupervision} or human annotation.
Distant supervision is a form of weak labeling where the labels are created automatically with a set of heuristics. These heuristics do not guarantee perfect labels, leading to noisy data that not only affects the training of models, but also leads to biased estimates of the models' performance. However, this process is fast and relatively cost efficient.
\\
Human annotation is an effective way to perform strong supervision. Although this reduces compounding of errors for downstream tasks, the obvious drawbacks are the marked increase in time and cost. These become prohibitively large when constructing larger datasets that can effectively train high-capacity models that can generalize to a variety of domains eg: \citet{transformers, gated_conv_lang_models}.
\\
Our contributions are:
\begin{itemize}
\item We introduce \webred{} - a diverse dataset for relation extraction. The text comes from a variety of publicly available sources on the internet that offer a multitude of domains and writing styles. We describe methods to collect $\sim$200M weakly supervised examples that can be used for supervised pre-training, and release $\sim$110K human annotated examples that allow us to fine-tune or train models and reliably evaluate their performance.
\item We show that pre-training relation extraction models on weakly-supervised data followed by fine-tuning on strongly-supervised data leads to models with higher F1-scores for relation extraction (see Table \ref{tab:hfive_results}).
\item We analyze the effects of data availability and quality (especially weak versus strong supervision) and stress on the importance of strong labels (Section \ref{sec:experiments_analysis}).
\end{itemize}
Relation extraction consists of many tasks: entity recognition, co-reference resolution, entity linking and then `slot-filling' which fills in the relations between the entities found in the text. Our dataset focuses on the task of `slot-filling' relations as a multi-class classification problem. Each example in our dataset consists of a single sentence from a web document whose entities are tagged to make the task easier, as in this example: ``\#\{SUBJ\}Alice lives in \#\{OBJ\}Baltimore''. This sentence is paired with a label that is a relation type (\textit{lives-in} for the preceding example) which is one from a pre-defined subset of WikiData\citep{wikidata} properties or `no relation' (P0, which denotes that the entities are not related). Section \ref{sec:post_processing} further explains how we chose the subset of relation types from WikiData. Every example in our dataset is thus a pair of \textit{(tagged-sentence, relation)}. Although a sentence may contain more than one unique entity pair, a \textit{tagged-sentence} is always unique because only the entity pair in the fact is tagged. Table \ref{tab:example_sentence} showcases a case of how examples are generated in our dataset. 
\\
To further advance the research and applications of relation extraction, we also release this dataset at \url{https://github.com/google-research-datasets/WebRED}.

\begin{table*}[h]
\begin{tabular}{l|l}
\textbf{Input Sentence} & Alice lives in Baltimore, and is married to Charlie.                                                                                                                   \\ \hline
\textbf{Example 1}      & \begin{tabular}[c]{@{}l@{}}\textbf{Tagged sentence}: \#\{SUBJ\}Alice lives in \#\{OBJ\}Baltimore, and is married to Charlie\\ \textbf{Label}: P551 (lives-in)\end{tabular}    \\ \hline
\textbf{Example 2}      & \begin{tabular}[c]{@{}l@{}}\textbf{Tagged sentence}: \#\{SUBJ\}Alice lives in Baltimore, and is married to \#\{OBJ\}Charlie\\ \textbf{Label}: P26 (spouse)\end{tabular}      \\ \hline
\textbf{Example 3}      & \begin{tabular}[c]{@{}l@{}}\textbf{Tagged sentence}: Alice lives in \#\{OBJ\}Baltimore, and is married to \#\{SUBJ\}Charlie\\ \textbf{Label}: P0 (no-relation)\end{tabular} \\ 
\end{tabular}
\caption{This showcases the kinds of examples that are generated from a single sentence in our dataset. Each example is a pair of a tagged-sentence and a relation type label.}
\label{tab:example_sentence}
\end{table*}

The rest of the paper is structured as follows:
\begin{enumerate}
    \item We highlight a few relation extraction datasets and other related work in Section \ref{sec:related_work}.
    \item In Section \ref{sec:dataset}, we elaborate on the methods used to construct \webred{}, the exact methods used for post-processing and filtering (Section \ref{sec:post_processing}) and describe its properties in Section \ref{sec:stratification}. We showcase the shortcomings of distant supervision and stress on the importance of human annotation in Section \ref{sec:result_human_annotation}.
    \item We present our pre-training and fine-tuning techniques used to train our models and show empirical results on our dataset in Section \ref{sec:experiments}. This section also describes in detail the two classes of models we use for relation classification, Transformers\citep{transformers} and BERT-style\citep{bert} Transformers. Further, it contains details of the task and all experimental settings used in this paper for reproducibility. Further, we also analyze the performance of our models specific to different settings in \ref{sec:experiments_analysis} and show that a combination of pre-training on a large weakly supervised subset and fine-tuning on human annotated data leads to the best performance.
    \item Finally, we conclude with Section \ref{sec:conclusion} with a discussion of our paper in context to existing work.
\end{enumerate}

\section{Related work} \label{sec:related_work}
There are many approaches to extracting relations between named entities \cite{classic_ner}, and several of them are detailed in \citet{rel_extract_survey_17} and  \citet{rel_extract_review_11}. In this paper, we focus on a supervised way of extracting relationships, that are one from a pre-defined set, between a pair of entities from sentences containing them. 
\\
Distant supervision \citep{distantsupervision} is widely used to collect data to learn structured information from unstructured data, and \citet{rel_ext_dist_supervision_survey} details some of the approaches and challenges of using it in the context of relation extraction. While there has been some work like \citet{improving_distant_supervision_label_propagation, improving_distant_supervision_inference_learning} that propose ways to improve distant supervision, we instead construct a large strongly-supervised dataset that in combination with a weakly-supervised dataset leads to training better relation extraction models (see Section \ref{sec:experiments_analysis}).
\\
One such dataset that is constructed with distant supervision is WikiFact\citep{wikifact_neural_klm}. It constructs examples by finding sentences in Wikipedia that contain mentions of the subject and object entities from WikiData\citep{wikidata} facts. However, this dataset is restricted to only the lead-section of the `film actor' subcategory in Wikipedia. The Wikidata/Wikipedia corpus introduced in \citet{factualaccuracy} extends this to facts from whole Wikipedia articles and contains several more categories. Wikipedia's writing style is constrained and models trained on this domain may not generalize to all types of text. Similarly, the Freebase/NYT\citep{freebase_nyt} dataset aligns Wikidata facts with text from NY Times articles.
\\
The TAC Relation Extraction Dataset \citep{tacred} is a strongly supervised dataset with 106,264 examples for 42 relation categories and is built on the TAC KBP\footnote{\url{https://tac.nist.gov/2017/KBP/}} corpus. However, our strongly supervised subset contains 111,717 examples with 523 relation categories, and contains more diverse forms of text. DocRED \citep{docred} is another dataset that contains a combination of weakly supervised and strongly supervised examples. However, it is built using only Wikipedia text and contains 63,427 human annotated examples and 1,508,320 examples constructed with distant supervision, compared to the $\sim$200M examples we collected for pre-training. 

Recently, pre-training \citep{layerwise_pretraining, why_pretraining_helps} has been used effectively to train higher capacity neural networks for language modeling \citep{bert, improving_lm_generative_pretraining} and relation extraction \citep{spanbert,simple_bert}. In this work, we compare using BERT-style \citep{bert} pre-training tasks against pre-training with the relation extraction task on our dataset and show that our method leads to better performance.

\section{Dataset} \label{sec:dataset}
In this section, we describe how we constructed the \webred{} dataset. Firstly, we collect a large weakly-supervised subset that can be used for pre-training and then select a subset of that for strongly-supervision via human annotation for fine-tuning and evaluation. The process to construct each part is described in detail in \ref{sec:distant_supervision} and \ref{sec:human_annotation}. To sample text from a variety of categories and writing styles, we surveyed a group of 10 human annotators to select web-domains that typically publish high linguistic quality and factually accurate content. We sampled web-pages from these domains and that formed the text corpus for \webred{}. For more details on how the text corpus for \webred{} was constructed please refer to Appendix \ref{appendix:web_domains}).
\subsection{Distant supervision} \label{sec:distant_supervision}
We make use of distant supervision \citep{distantsupervision} to collect our weakly supervised pre-training data. We perform Named Entity Recognition (NER) and Co-reference Resolution (CoRef) on every document in our text corpus\footnote{We make use of a proprietary NER and CoRef system and release the results as part of our dataset. However, there are publicly available alternatives such as: \url{https://stanfordnlp.github.io/CoreNLP/}, \url{https://github.com/huggingface/neuralcoref}.}. 
If there are two or more entities in a sentence from these documents, we try finding a WikiData \citep{wikidata} fact tuple (\textit{subject, relation, object}) that contains a pair of unique entities as \textit{subject} and \textit{object}. If such a tuple is found, this sentence is marked as a positive match for the relation. If the sentence does not match any fact tuple, it is marked as containing no relation (\textit{P0}). This means that our dataset contains all facts among all the entity pairs found in the document that are also in WikiData. We restricted our text corpus and WikiData facts to English-language content and fact tuples.

\subsection{Human annotation} \label{sec:human_annotation}
\begin{figure*}[t!]
  \begin{center}
    \centering
    \includegraphics[width=\linewidth]{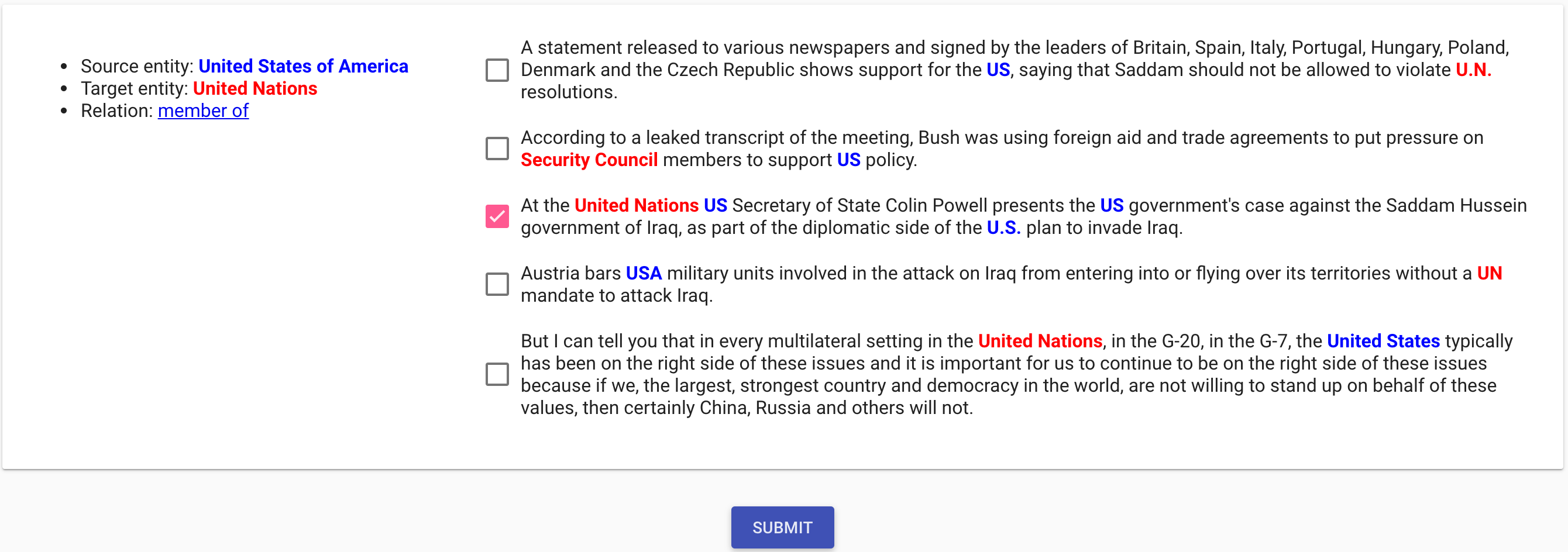}%
    \caption{This figure shows the UI presented to human annotators. They are asked to \textit{deselect} from a list of sentences possibly expressing the fact shown on the left, if the sentence does not express the relation between the subject (source entity) and object (target entity). In this case, only one sentence is selected as expressing the relation.}
    \label{fig:human_annotation_gui}
  \end{center}
\end{figure*}

We make use of crowd-sourcing to strongly-supervise a subset of our weakly-supervised data (Section \ref{sec:distant_supervision}). This subset is created by choosing a uniform sample of documents from those in our weakly-supervised subset. For each fact tuple that matched a sentence in these documents, annotators were asked to deselect all sentences that did not express the relation. Figure \ref{fig:human_annotation_gui} shows an example. Sentences that were deselected were labeled as \textit{P0} (no relation) for the given entity pair. Annotators were instructed to not use any external knowledge and to only assess whether the sentence directly implies the relation between the entities. Further information about human annotation can be found in Appendix \ref{appendix:annotation}.\\
Our strongly supervised data was labeled in two ways:
\begin{enumerate}
    \item A large subset was initially labeled by 2 annotators. If they disagreed on the labels of an example, another annotator labeled the same example. The final label was the majority of the 3 votes. We call this subset \htwentyone. 90\% of this subset is used for training or fine-tuning and 10\% of it is used as part of the test set. We found that only $\sim$20\% of this subset required arbitration by a third annotator.
    \item A smaller subset was labeled by 5 annotators. Once again, the final label was the majority of the 5 votes. This subset is entirely used as part of the test set, and we refer to it as \hfive.
\end{enumerate}

Since knowledge bases are not complete, a sentence labeled as P0 with this process may express a different relation that currently has no fact tuple in WikiData. However, to make the task easier and faster to complete, we do not ask annotators to re-label negative examples due to the complexity of selecting a new label from hundreds of relations. 

\subsection{Post-processing} \label{sec:post_processing}
Besides collecting data as described in the previous sections, we describe further procedures we used to construct our dataset. Firstly, on inspecting the examples annotated by humans (Section \ref{sec:human_annotation}), we found that 605 out of the 1,027 relation types we considered had no positive labels. This is due to reasons that include many WikiData relation types like ID numbers are not expressed in natural language. All examples containing these relation types were then removed from all subsets of our dataset.\\
With distant supervision (Section \ref{sec:distant_supervision}), we collected $\sim$500M examples. From this set, we used rejection sampling\citep{rejection_sampling} to collect $\sim$200M examples such that the distribution of examples per relation type in this subset that we call \ds{} matches that of \htwentyone{}. 

\subsection{Dataset stratification} \label{sec:stratification}
After data collection and filtering (Section \ref{sec:post_processing}), our dataset is comprised of 199,786,781 examples in \ds (weakly supervised) and 117,717 examples in \htwentyone (107,819) + \hfive (3898) with 523 relation types and 65\% of our dataset being negative examples (the relation type is P0). We compare this with some other datasets in Table \ref{tab:dataset_comparison}. Figure (Appendix) \ref{fig:appendix_rel_dist} shows the distribution of the number of examples available in our dataset per relation types.
\ds forms our pre-training subset, while 90\% of \htwentyone is used for fine-tuning. All of \hfive and 10\% of \htwentyone are set aside as our test subset. While training, 10\% of the corresponding subset (i.e \ds or \htwentyone) is set aside for cross-validation.


\begin{table}[t]
\addtolength{\tabcolsep}{-3pt}
\begin{tabular}{p{3.2cm}|p{0.8cm}|p{2cm}|p{1cm}}
\textbf{Dataset}                                    & \textbf{\#Rel}  & \textbf{\#Examples}     & \textbf{\%Neg}  \\ \hline
\makecell[l]{TACRED}                          & 42     & 106,264     & 79.5   \\ \hline
\makecell[l]{DocRED\\(human-annotated)}       & 96     & 63,427      & N/A    \\ \hline
\makecell[l]{DocRED\\(weakly-supervised)}     & 96     & 1,508,320   & N/A    \\ \hline
\makecell[l]{\webred{}\\(human-annotated)}    & 523    & 117,717     & 65     \\ \hline
\makecell[l]{\webred{}*\\(weakly-supervised)}  & 523    & 199,786,781 & 65     \\ \hline
\end{tabular}
\caption{A comparison of existing datasets and our proposed \webred{} dataset. \#Rel denotes the number of relation types and \%Neg is the percentage of negative examples in the dataset. *Note that we only describe the method to collect our pre-training set and release the human-annotated training data.}
\label{tab:dataset_comparison}
\end{table}

\subsection{Results of human annotation} \label{sec:result_human_annotation}
Distant supervision is a heuristic-based labeling mechanism that is bound to lead to noisy labels. Table \ref{tab:entity_type_ds_acc} shows the true accuracy of distant supervision on our dataset on a few types of entities. We find that by looking at the labels assigned to the \htwentyone subset before human annotation (i.e with distant supervision) and comparing them to the true labels. This result suggests that distant supervision may work with reasonable accuracy for relation types like \textit{date of birth}, \textit{year of establishment} etc, that connect to `Time' based entities but is prohibitively inaccurate for relation types like \textit{distance to}, \textit{weight of} etc that connect to `Quantity' based entities.\\
Similarly, Table \ref{tab:rel_type_ds_acc} shows the accuracy of distant supervision for a few relation types (we picked a small subset of the most frequent relation types that showcase different behaviors). It is apparent from this that human annotation changes the underlying input distribution for a few relation types, stressing on the importance of strong supervision for training accurate models.\\

\begin{table}[h]
\begin{tabular}{l|l}
\textbf{Entity Type} & \textbf{DS Accuracy} \\ \hline
Quantity             & 0.009                \\ 
Time                 & 0.823                \\ 
Others               & 0.462                \\ 
\end{tabular}
\caption{True accuracy of the labels generated by distant supervision (DS) per the type of entity that is found by human annotation. \textit{Others} includes persons, organizations, locations, products etc.}
\label{tab:entity_type_ds_acc}
\end{table}

\begin{table}[t]
\begin{tabular}{p{1cm}|p{3cm}|p{1cm}|p{1cm}}
\textbf{Prop} & \textbf{Relation}                                & \textbf{Freq} & \textbf{DS Acc} \\ \hline
P17               & country                                          & 39151              & 0.847                \\ 
P530              & diplomatic relation                              & 29901              & 0.011                \\ 
P131              & \makecell[l]{located in the \\administrative\\territorial entity} & 16268              & 0.880                \\ 
P47               & shares border with                               & 14895              & 0.020                \\ 
P36               & capital                                          & 4373               & 0.023                \\ 
P54               & member of sports team                            & 2251               & 0.842                \\ 
P26               & spouse                                           & 1113               & 0.363                \\ 
P569              & date of birth                                    & 632                & 0.976                \\ 
P571              & inception                                        & 435                & 0.63                 \\ 
P138              & named after                                      & 360                & 0.105                \\ 
\end{tabular}
\caption{Accuracy of distant supervision (\textit{DS Acc}) per relation type. \textit{Prop} is the WikiData property corresponding to the relation, and \textit{Freq} is the number of occurrences of the relation in \htwentyone .}
\label{tab:rel_type_ds_acc}
\end{table}

\section{Experimental setup} \label{sec:experiments}
We treat relation extraction as a multi-class classification problem where models are trained with the Cross-Entropy loss function to pick one relation. The inputs to these models are individual sentences where we added hints by tagging the subject and object entities with special tokens as in this example: ``\textit{\#SUBJ\{Alice\} grew up in the town of \#OBJ\{Baltimore\}}''. Labels are presented to the model as the one-hot encoding of the true labels. Our relation extraction classification models are based on the Transformer\citep{transformers} architecture. These models consist of 1 embedding layer followed by a series of Transformer encoder layers that feed into a Softmax classification layer (which is a fully-connected dense layer followed by the Softmax operation). To study the effect of pre-training these models, we consider the scenarios below. The training, validation and test set stratification are described in Section \ref{sec:stratification}. An important point of note, is that while we are unable to release the full set of training data that we collected in this process due to potential copyright issues, we release all data that are in open domains. We report numbers below on the full-sized dataset we collected internally (this contains 173,140 human-supervised examples, and is a super-set of the released version). Additionally, we only selected 420 relations from our human annotated data that contained at least more than one training instance. 
\begin{enumerate}
    \item Training a model for relation extraction only using data from \htwentyone. We use a Transformer-based classifier with 6 encoder layers, hidden-size of 512 and 8 attention heads. We henceforth refer to this setting as $T_{base}$. This model is trained using the AdaFactor \citep{adafactor} optimizer with a learning rate of 1e-2 and batch size of 64 for 50,000 steps.
    \item Pre-training for relation extraction on \ds, and then fine-tuning on \htwentyone. We use the $T_{base}$ model and pre-train it with a learning rate of 1e-2 and batch size of 8192 for 500,000 steps. We then fine-tune it for 3,000 steps with learning rate of 1e-3 and batch size of 256. We use the AdaFactor optimizer for both pre-training and fine-tuning, and did not reset any optimizer variables (momentum/velocity) for fine-tuning.
    \item Using a BERT \citep{bert} language model (we use the Large-Cased(Original) model released in \url{https://github.com/google-research/bert}) that was pre-trained on BooksCorpus\citep{bookscorpus} and English Wikipedia, that is then fine-tuned for relation extraction on \htwentyone. We append a Softmax classification layer on top of the language model to fine-tune it for relation extraction. We use a Transformer-based classifier with 24 encoder-decoder layers, hidden-size of 1024, 16 attention heads and use the GeLU \citep{gelu} activation. This model was fine-tuned using the AdaFactor optimizer with a learning rate of 1e-5 and batch size of 32 for 20,000 steps.
\end{enumerate}
All the models described above use a maximum input length of 128 and use sub-word tokenization \citep{subword_tokenizer} to encode input text. This means that sentences with more than 128 tokens are truncated before being processed by our models. The choice of parameters described above were a result of tuning hyperparameters and early-stopping on the validation set. As described above, we use a 10\% of \ds for validation during pre-training, and then change it to 10\% of \htwentyone for validation during fine-tuning or for training on it from scratch.\\

Table \ref{tab:hfive_results} presents the results of the above scenarios as the performance on our test set. The performance of our models is presented in terms of Precision, Recall and F1 as defined by \citet{tacred}.  We discuss these results in the next Section (\ref{sec:experiments_analysis}).
\begin{table*}[t]
\begin{tabular}{l|l|l|l|lll|l}
\textbf{\makecell[l]{Pre-training/\\Training}} & \textbf{\makecell[l]{Pre-training\\task}} & \textbf{Fine-tuning} & \textbf{Model}      & \textbf{P}    & \textbf{R}    & \textbf{F1*}   & F1(P0) \\ \hline
\htwentyone                    & RE                         & -                    & $T_{base}$  & 0.31          & 0.24          & 0.27          & 0.81            \\ \hline
\makecell[l]{BERT\\(Books + Wikipedia)}       & \makecell[l]{BERT\\Masked LM}             & \htwentyone          & $T_{large}$ & 0.56          & 0.50          & 0.53          & 0.87            \\ \hline
\ds                            & RE                         & -                    & $T_{base}$  & 0.28          & \textbf{0.81} & 0.42          & 0.16            \\ \hline
\ds                            & RE                         & \htwentyone          & $T_{base}$  & \textbf{0.64} & 0.69          & \underline{\textbf{0.67}} & 0.88   \\ \hline
\end{tabular}
\caption{Model performance on the test set (Section \ref{sec:stratification}). \textit{Pre-training/Training} is the data used to train the model on the task specified under \textit{Pre-training task} (where $RE$ is relation extraction, and \textit{BERT Masked LM} is from \citet{bert}). $T_{large}$ and $T_{base}$ are Transformer models described above. $P$, $R$, $F1$ and $F1(P0)$ denote Precision, Recall, F1* and F1 for the `no-relation' relation type on our test set.}
\label{tab:hfive_results}
\end{table*}

\subsection{Analysis} \label{sec:experiments_analysis}
\begin{figure*}[t]
  \centering
  \subfloat[Precision]{\includegraphics[width=0.4\textwidth]{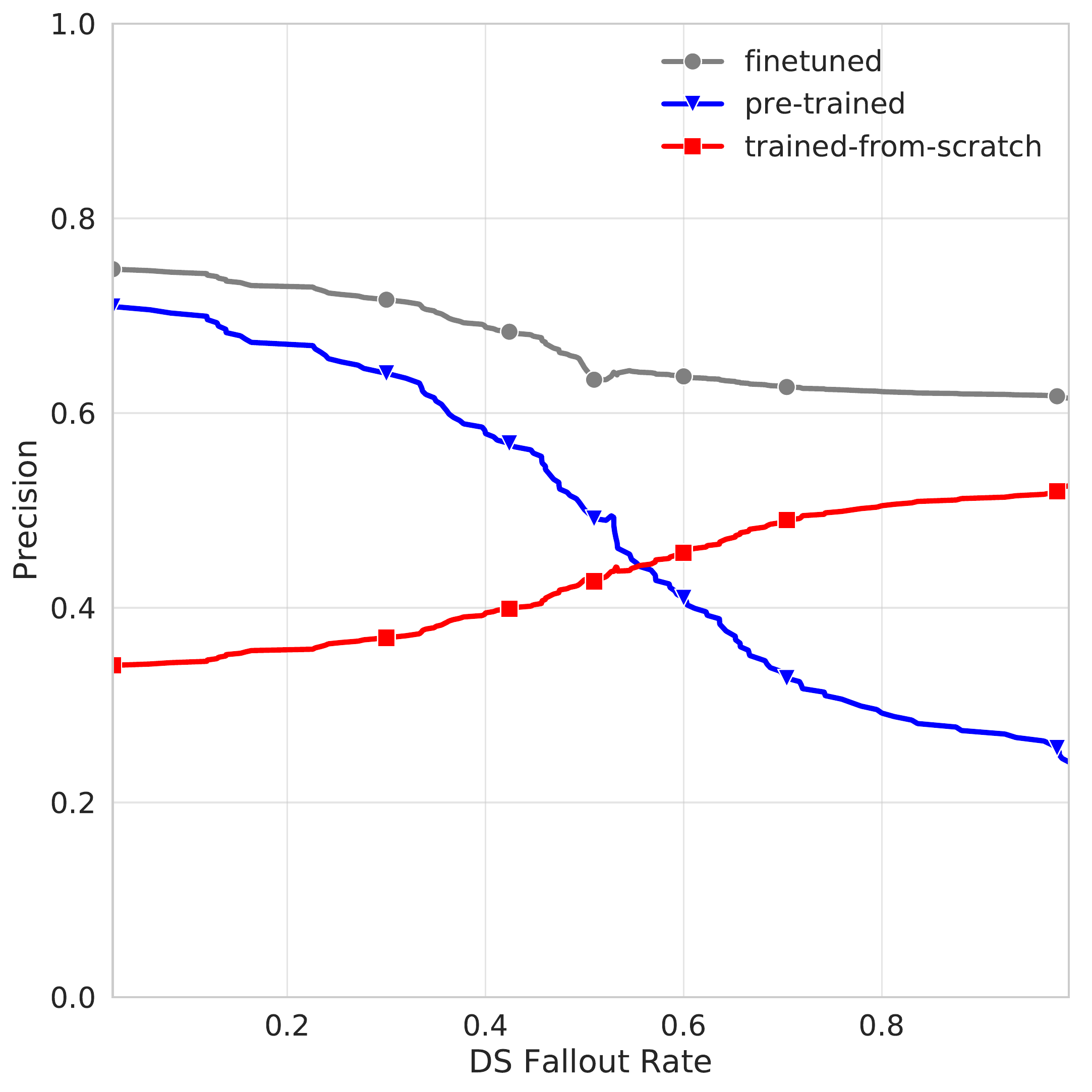}\label{fig:webfact_ds_fallout_prec}}%
   \centering
  \subfloat[Recall]{\includegraphics[width=0.4\textwidth]{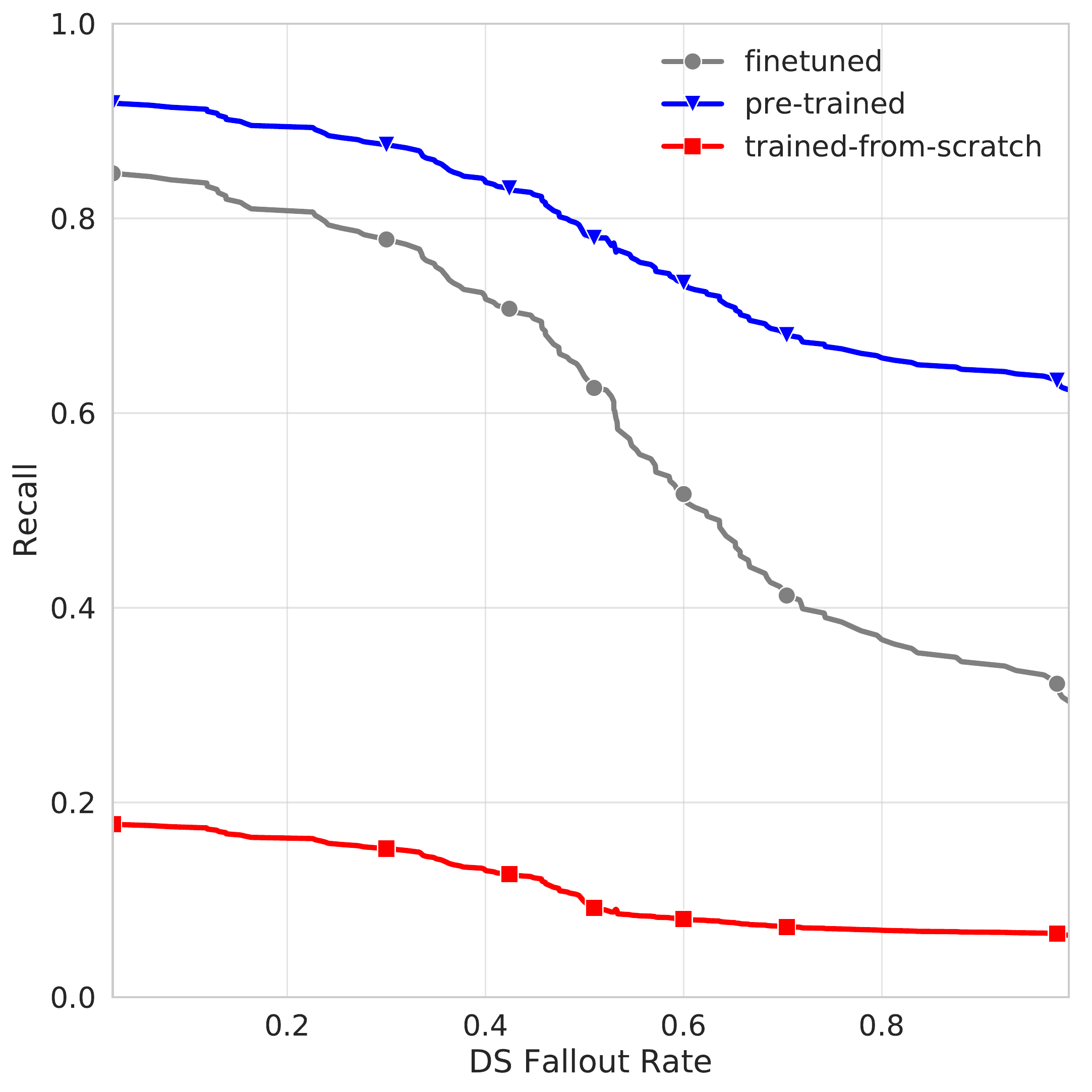}\label{fig:webfact_ds_fallout_recall}}%
  \caption{Performance of $T_{base}$ (Section \ref{sec:experiments}) that is only trained on \htwentyone (\textit{trained-from-scratch}), pre-trained on \ds (\textit{pre-trained}) and fine-tuned on \htwentyone(\textit{finetuned}) across relations and their fallout rate of distant supervision. Fallout rate is a measure of how many labels assigned by distant supervision changed after human annotation (i.e $1 - accuracy$)}
  \label{fig:webfact_ds_fallout_rate}
\end{figure*}

From Table \ref{tab:hfive_results}, we observe that all models perform better with pre-training. We hypothesize that \htwentyone (strongly-supervised) alone does not form a big enough dataset to train high capacity models that can generalize well. 
Although BERT-style masked language model training helps, we find that the best pre-training task is relation extraction with weakly supervised labels. This is despite the shift in label distribution after human annotation as discussed in Section \ref{sec:result_human_annotation}.\\
In the labeling process as described in Section \ref{sec:human_annotation}, examples with P0s are only created by deselecting sentences that do not express a specific relation, and this leads to strong supervision for P0s with a variety of text. This is unlike distant supervision, where examples for P0s are created if a fact tuple between arbitrary pairs of recognized entities is not found in the knowledge base, which leads to noisy P0 examples. As seen in Table \ref{tab:hfive_results}, models that are not trained or fine-tuned on \htwentyone perform poorly on the P0 relation type on our strongly supervised test set, suggesting that the distribution of examples for P0 with distant supervision is widely different from strong supervision.\\
We also observe from Table \ref{tab:perf_after_finetuning} that the performance (precision/recall) of models across some relations (Appendix \ref{appendix:model_perf} contains a more complete version of the same table) changes depending on how it was trained. Models that are pre-trained and then fine-tuned on \htwentyone learn to trade recall for precision for relations with high fallout rate (where fallout is $1 - accuracy$ of the labels assigned by distant supervision) as they learn from stronger supervision on negative examples, while also consistently outperforming the model only trained on \htwentyone indicating that the model benefits from pre-training.\\
We similarly see from Figures \ref{fig:webfact_ds_fallout_prec} and \ref{fig:webfact_ds_fallout_recall} that the fallout rate of distant supervision (where fallout is $1 - accuracy$ of the labels assigned by distant supervision) affects the performance per relation. Human annotation distinguishes false positives from true positives on a subset of the distantly supervised data, creating `stronger' negative examples than those synthetically generated by randomly pairing uncorrelated entities within a given sentence. The model fine-tuned on human annotated data has a better F1 score driven by significantly higher precision and slightly lower recall. Since the model that is trained only on \htwentyone never sees weak labels, its precision does not decrease with the fallout rate. However, since it sees considerably smaller amounts of data, it has low recall and never outperforms the pre-trained model that is fine-tuned on \htwentyone.

\begin{table*}[t!]
    \centering
    \begin{tabular}{p{3.8cm}|p{1cm}p{1cm}|p{1cm}p{1cm}|p{1cm}p{1cm}|p{0.8cm}}
    \makecell[l]{\footnotesize{\textbf{Property}}} & \multicolumn{2}{l|}{\makecell[l]{\footnotesize{\textbf{Scratch}}}} & \multicolumn{2}{l|}{\makecell[l]{\footnotesize{\textbf{Pretrained}}}} & \multicolumn{2}{l|}{\makecell[l]{\footnotesize{\textbf{Finetuned}}}} & \footnotesize{\textbf{\makecell[l]{DS\\Fallout}}} \\ \hline
    & \textbf{P} & \textbf{R} & \textbf{P} & \textbf{R} & \textbf{P} & \textbf{R} &
    \\ \hline
        country & 0.33 & 0.36 & 0.58 & 0.63 & 0.81 & 0.58 & 0.15 \\ 
diplomatic relation & 0.33 & 0.17 & 0.01 & 0.53 & 1.00 & 0.00 & 0.99 \\ 
located in the administrative territorial entity & 0.35 & 0.23 & 0.44 & 0.96 & 0.57 & 0.79 & 0.12 \\ 
shares border with & 0.26 & 0.20 & 0.01 & 0.88 & 0.60 & 0.12 & 0.98 \\ 
capital & 0.22 & 0.18 & 0.01 & 0.27 & 0.71 & 0.45 & 0.98 \\ 
member of sports team & 0.45 & 0.42 & 0.73 & 0.97 & 0.76 & 0.92 & 0.16 \\ 
spouse & 0.47 & 0.49 & 0.38 & 0.94 & 0.71 & 0.86 & 0.64 \\ 
date of birth & 0.60 & 0.80 & 0.96 & 1.00 & 0.99 & 1.00 & 0.02 \\ 
director & 0.00 & 0.00 & 0.20 & 0.50 & 0.64 & 0.56 & 0.57 \\ 
    \end{tabular}
    \caption{Performance (Precision/Recall) of models on a few relation types (\textit{Property}) when they are trained only on \htwentyone (\textit{Scratch}), only \ds (\textit{Pretrained}) or pre-trained on \ds and then fine-tuned on \htwentyone (\textit{Finetuned}). \textit{DS Fallout} is $1 - accuracy$ of the labels assigned by distant supervision. The relations in this table are a subset of the top-25 most frequent relations in our dataset that also include those with high fallout rates.}
    \label{tab:perf_after_finetuning}
\end{table*}

\begin{figure}[h]
    \begin{tabular}{@{}c@{}}

  \includegraphics[width=0.9\linewidth]{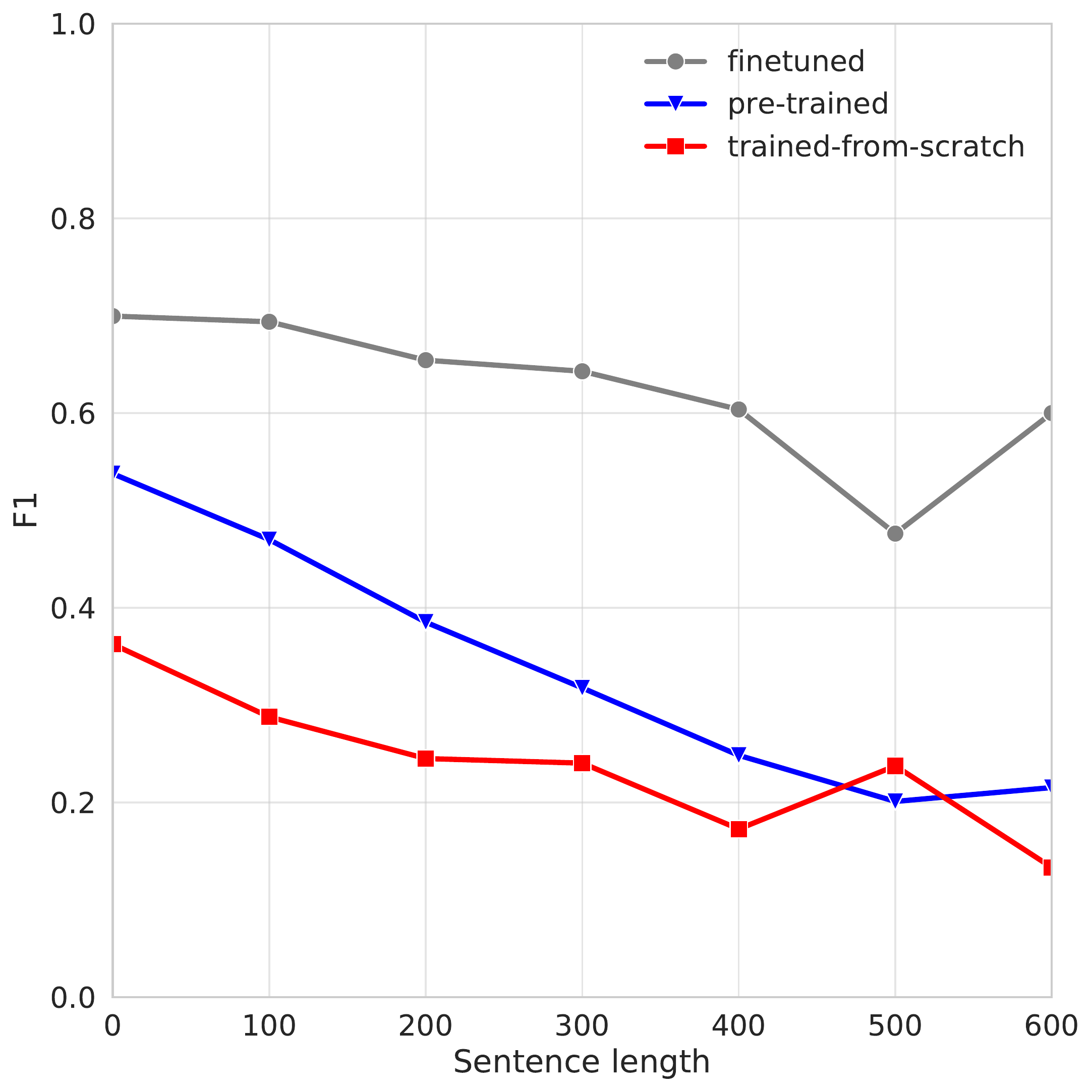}\label{fig:sentence_length_perf}\\[\abovecaptionskip]
    \small (a) F1 vs Sentence length%
\end{tabular}
 \begin{tabular}{@{}c@{}}
\includegraphics[width=0.9\linewidth]{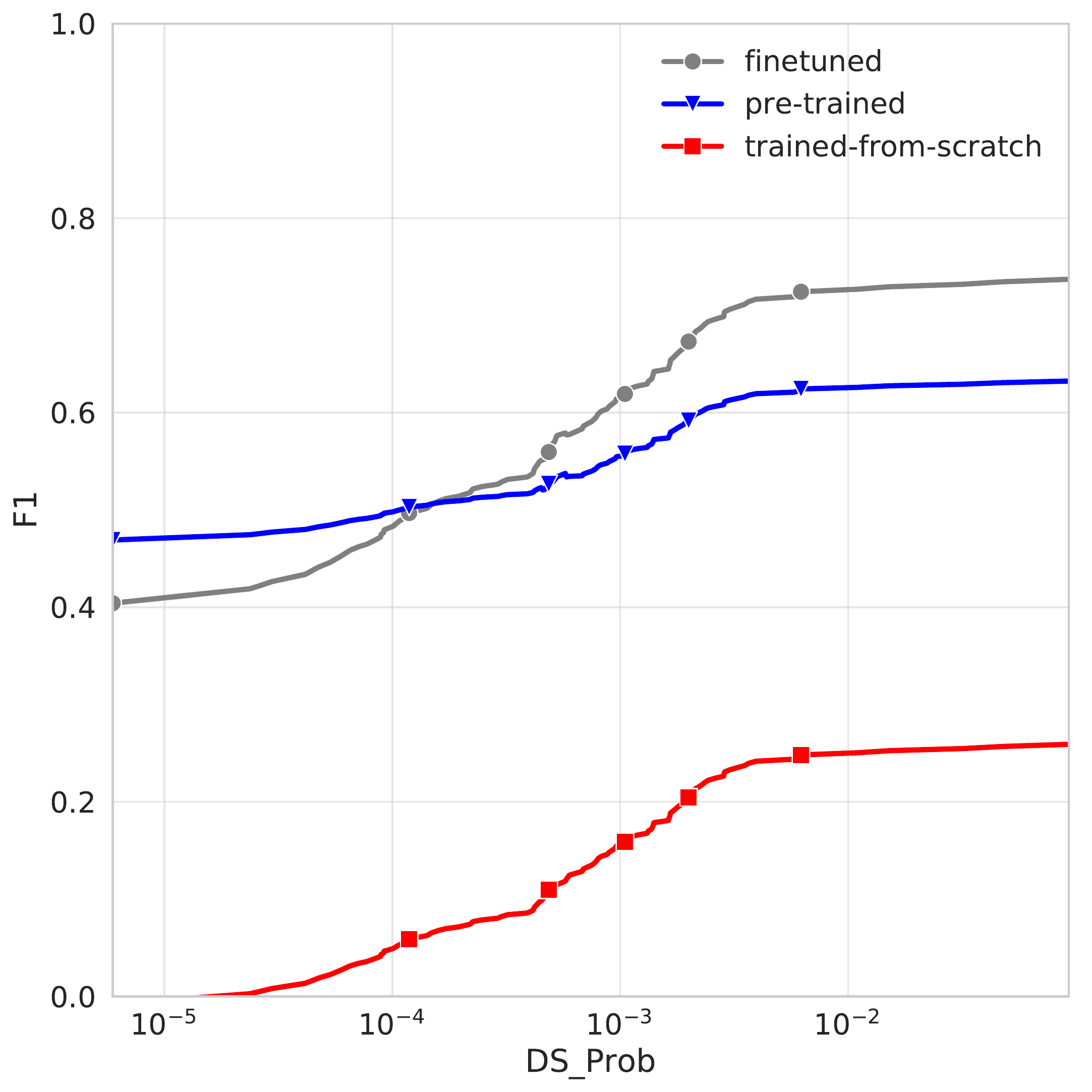}\label{fig:ds_prob_perf}\\[\abovecaptionskip]
    \small (b) F1 vs Frequency of relation type%
\end{tabular}
  \caption{Performance (F1) of $T_{base}$ (Section \ref{sec:experiments}) that is only trained on \htwentyone (\textit{trained-from-scratch}), pre-trained on \ds (\textit{pre-trained}) and fine-tuned on \htwentyone(\textit{finetuned}). (a) shows the F1-score across sentence lengths and (b) shows the F1-score across relations and their probability of occurrence in our pre-training set \ds.}
  \label{fig:sent_length_ds_prob}
\end{figure}

Figure \ref{fig:sent_length_ds_prob}(a) shows the performance (F1) of a model on our test set, grouped by length of the input sentence. We can see that model performance decreases with increasing sentence lengths. However, pre-training combined with fine-tuning helps the most in improving performance for all different sentence lengths.\\

A somewhat obvious result is presented in Figure \ref{fig:sent_length_ds_prob}(b) which indicates that increased availability of examples helps improve the performance for relation types, however there is a point beyond which the gains are minimal. This implies that balancing the frequency of positive examples across all relations is paramount for good overall performance. 


\section{Conclusion} \label{sec:conclusion}
We introduce and release \webred, a large and diverse human annotated relation extraction dataset that enables training high capacity models for extracting relations from text found on the web. With the methods we describe for collecting pre-training data, it offers a variety of writing styles and domains of text. \\
We also presented an analysis on the shortcomings of distant supervision for this task by comparing it against human annotations, along with the change in performance of our models depending on the availability of data per relation types and the labeling accuracy of distant supervision.\\
In summary. we show that pre-training models with weakly-supervised data followed by fine-tuning on smaller strongly-supervised data is cost effective and leads to better relation extraction performance.
Finally, we release our dataset at \url{https://github.com/google-research-datasets/WebRED}.

\clearpage
\bibliographystyle{acl_natbib}
\bibliography{main}

\clearpage
\appendix
\section*{Appendix}
\section{Model Performance} \label{appendix:model_perf}
Table \ref{tab:appendix_model_perf} details the performance (Precision and Recall) of our model across several relation types in our dataset. They are ordered by their frequency of occurrence in our weakly-supervised dataset \ds. 

\begin{table*}
\centering
    \begin{tabular}{|p{3.8cm}|p{1cm}|p{1cm}|p{1cm}|p{1cm}|p{1cm}|p{1cm}|p{0.8cm}|}
        \hline
\makecell[l]{\footnotesize{\textbf{Property}}} & \multicolumn{2}{|l|}{\makecell[l]{\footnotesize{\textbf{Scratch}}}} & \multicolumn{2}{|l|}{\makecell[l]{\footnotesize{\textbf{Pretrained}}}} & \multicolumn{2}{|l|}{\makecell[l]{\footnotesize{\textbf{Finetuned}}}} & \footnotesize{\textbf{\makecell[l]{DS\\Fallout}}} \\ \hline
 & \textbf{P} & \textbf{R} & \textbf{P} & \textbf{R} & \textbf{P} & \textbf{R} &
    \\ \hline
country & 0.33 & 0.36 & 0.58 & 0.63 & 0.81 & 0.58 & 0.15 \\ \hline
diplomatic relation & 0.33 & 0.17 & 0.01 & 0.53 & 1.00 & 0.00 & 0.99 \\ \hline
located in the administrative territorial entity & 0.35 & 0.23 & 0.44 & 0.96 & 0.57 & 0.79 & 0.12 \\ \hline
shares border with & 0.26 & 0.20 & 0.01 & 0.88 & 0.60 & 0.12 & 0.98 \\ \hline
contains administrative territorial entity & 0.42 & 0.31 & 0.49 & 0.99 & 0.59 & 0.84 & 0.12 \\ \hline
country of citizenship & 0.27 & 0.23 & 0.40 & 0.92 & 0.74 & 0.75 & 0.53 \\ \hline
capital & 0.22 & 0.18 & 0.01 & 0.27 & 0.71 & 0.45 & 0.98 \\ \hline
capital of & 0.33 & 0.17 & 0.03 & 0.17 & 1.00 & 0.08 & 0.98 \\ \hline
encodes & 0.45 & 0.14 & 0.00 & 0.00 & 0.38 & 0.17 & 0.80 \\ \hline
member of sports team & 0.45 & 0.42 & 0.73 & 0.97 & 0.76 & 0.92 & 0.16 \\ \hline
has part & 0.23 & 0.09 & 0.58 & 0.80 & 0.76 & 0.72 & 0.51 \\ \hline
continent & 0.29 & 0.31 & 0.32 & 0.91 & 0.46 & 0.71 & 0.66 \\ \hline
part of & 0.15 & 0.11 & 0.49 & 0.74 & 0.62 & 0.79 & 0.42 \\ \hline
member of political party & 0.41 & 0.32 & 0.42 & 0.95 & 0.49 & 0.84 & 0.36 \\ \hline
spouse & 0.47 & 0.49 & 0.38 & 0.94 & 0.71 & 0.86 & 0.64 \\ \hline
place of birth & 0.68 & 0.44 & 0.33 & 0.92 & 0.84 & 0.95 & 0.72 \\ \hline
member of & 0.20 & 0.29 & 0.60 & 0.85 & 0.85 & 0.85 & 0.43 \\ \hline
owned by & 0.10 & 0.08 & 0.30 & 0.85 & 0.51 & 0.50 & 0.48 \\ \hline
head of government & 0.14 & 0.16 & 0.43 & 0.88 & 0.55 & 0.96 & 0.55 \\ \hline
location of formation & 0.08 & 0.06 & 0.19 & 0.88 & 0.56 & 0.29 & 0.70 \\ \hline
employer & 0.15 & 0.06 & 0.60 & 0.91 & 0.67 & 0.87 & 0.40 \\ \hline
cast member & 0.26 & 0.12 & 0.60 & 0.83 & 0.83 & 0.79 & 0.25 \\ \hline
country of origin & 0.14 & 0.08 & 0.42 & 0.68 & 0.70 & 0.61 & 0.57 \\ \hline
performer & 0.20 & 0.13 & 0.59 & 0.91 & 0.68 & 0.89 & 0.34 \\ \hline
owner of & 0.33 & 0.12 & 0.35 & 0.96 & 0.54 & 0.76 & 0.46 \\ \hline
parent organization & 0.21 & 0.16 & 0.55 & 0.64 & 0.78 & 0.62 & 0.49 \\ \hline
official language & 0.00 & 0.00 & 0.06 & 1.00 & 0.00 & 0.00 & 0.88 \\ \hline
head of state & 0.02 & 0.17 & 0.24 & 0.96 & 0.54 & 0.57 & 0.60 \\ \hline
author & 0.32 & 0.19 & 0.54 & 0.87 & 0.69 & 0.67 & 0.47 \\ \hline
place of death & 0.00 & 0.00 & 0.33 & 0.81 & 0.79 & 0.94 & 0.83 \\ \hline
developer & 0.11 & 0.07 & 0.49 & 0.80 & 0.65 & 0.70 & 0.46 \\ \hline
subclass of & 0.12 & 0.04 & 0.31 & 0.96 & 0.54 & 0.60 & 0.70 \\ \hline
applies to jurisdiction & 0.14 & 0.04 & 0.47 & 0.69 & 0.48 & 0.49 & 0.55 \\ \hline
founded by & 0.00 & 0.00 & 0.32 & 0.44 & 0.75 & 0.50 & 0.72 \\ \hline
date of birth & 0.60 & 0.80 & 0.96 & 1.00 & 0.99 & 1.00 & 0.02 \\ \hline
parent taxon & 0.58 & 0.16 & 0.53 & 0.91 & 0.66 & 0.84 & 0.53 \\ \hline
followed by & 0.31 & 0.20 & 0.43 & 0.88 & 0.49 & 0.63 & 0.59 \\ \hline
subsidiary & 0.24 & 0.23 & 0.41 & 0.72 & 0.55 & 0.77 & 0.47 \\ \hline
follows & 0.27 & 0.21 & 0.50 & 0.92 & 0.52 & 0.63 & 0.53 \\ \hline
operator & 0.17 & 0.08 & 0.42 & 0.96 & 0.67 & 0.50 & 0.32 \\ \hline
league & 0.43 & 0.43 & 0.91 & 1.00 & 0.97 & 0.93 & 0.08 \\ \hline
educated at & 0.50 & 0.31 & 0.67 & 0.82 & 0.85 & 0.84 & 0.37 \\ \hline
location & 0.07 & 0.02 & 0.59 & 0.82 & 0.77 & 0.71 & 0.23 \\ \hline
\end{tabular}
\caption{Performance (Precision/Recall) of models on some of the relation types (\textit{Property}) when they are trained only on \htwentyone (\textit{Scratch}), only \ds (\textit{Pretrained}) or pre-trained on \ds and then fine-tuned on \htwentyone (\textit{Finetuned}). \textit{DS Fallout} is $1 - accuracy$ of the labels assigned by distant supervision.}
\label{tab:appendix_model_perf}
\end{table*}

\section{Human annotation} \label{appendix:annotation}
This section contains further information about the procedures used to guide human annotation of our strongly-supervised subset.
The human annotators had access to the annotation instructions shown in Table~\ref{tab:annotation_guide}.

\begin{table*}[t]
    \centering
    \begin{tabular}{|p{16cm}|}
        \hline
        \textbf{Annotation Guide}\\ \hline
        
          \begin{center}
            Before annotation:
            \includegraphics[width=\textwidth]{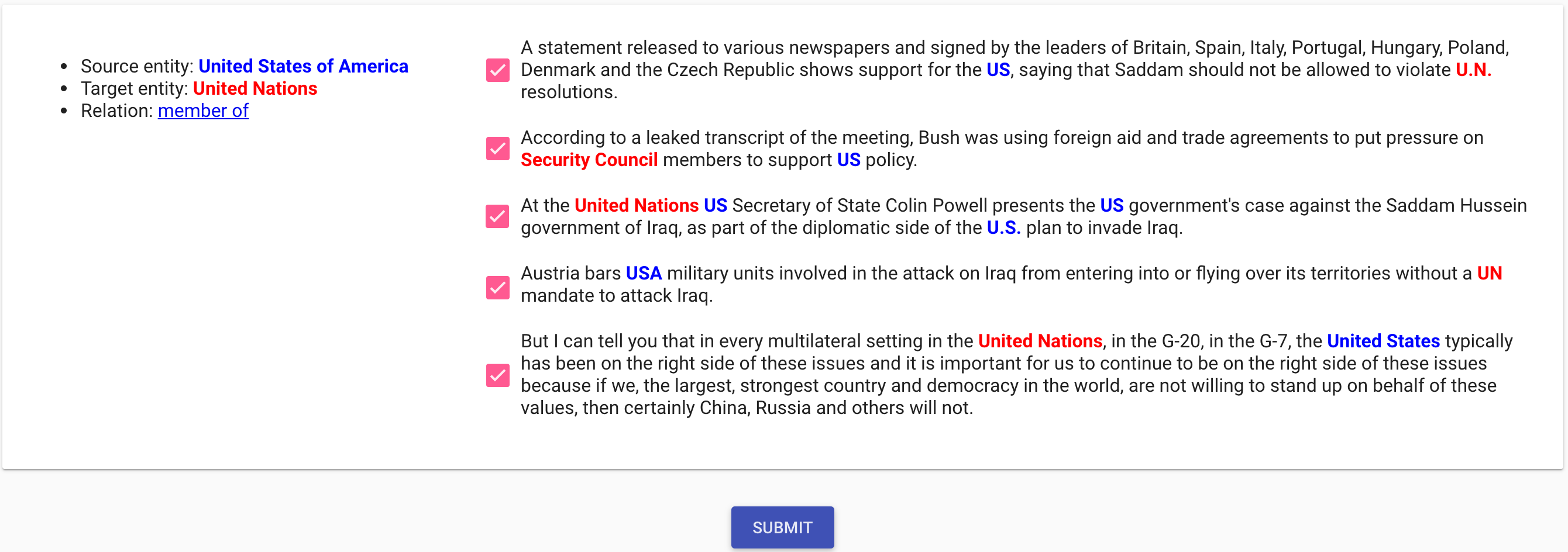}
            After annotation:
            \includegraphics[width=\textwidth]{annotation_gui_screenshot.png}
          \end{center}
        
        On the left hand side a fact (i.e. a directed relationship between the source and target entities) is shown. On the right hand side a list of sentences containing both the source and target entities are shown. In these sentences, color code is used to highlight the mentions of the source and target entities.\\
        \\
        Task: Deselect each sentence that does not explicitly express the given relation between the source and target entities. The relation has to be either directly stated or can be inferred from the sentence.\\
        \\
        Additional considerations: \\
        \begin{itemize}
        \item In some cases, a mention of an entity (i.e. color-coded phrase) in a sentence may be inaccurate. In these cases, please ignore this inaccuracy and assume that the mention refers to the highlighted entity on the left hand side.
        \item Do not look up external sources when answering questions. Rely on each sentence’s text and the linked relation definition alone.
        \end{itemize}\\
         \hline
    \end{tabular}
    \caption{The above annotation instructions were used to guide the annotators to perform human annotation.}
    \label{tab:annotation_guide}
\end{table*}

\section{Dataset} \label{appendix:dataset}
\begin{figure*}
  \centering
  \subfloat[Tacred]{\includegraphics[width=0.5\linewidth]{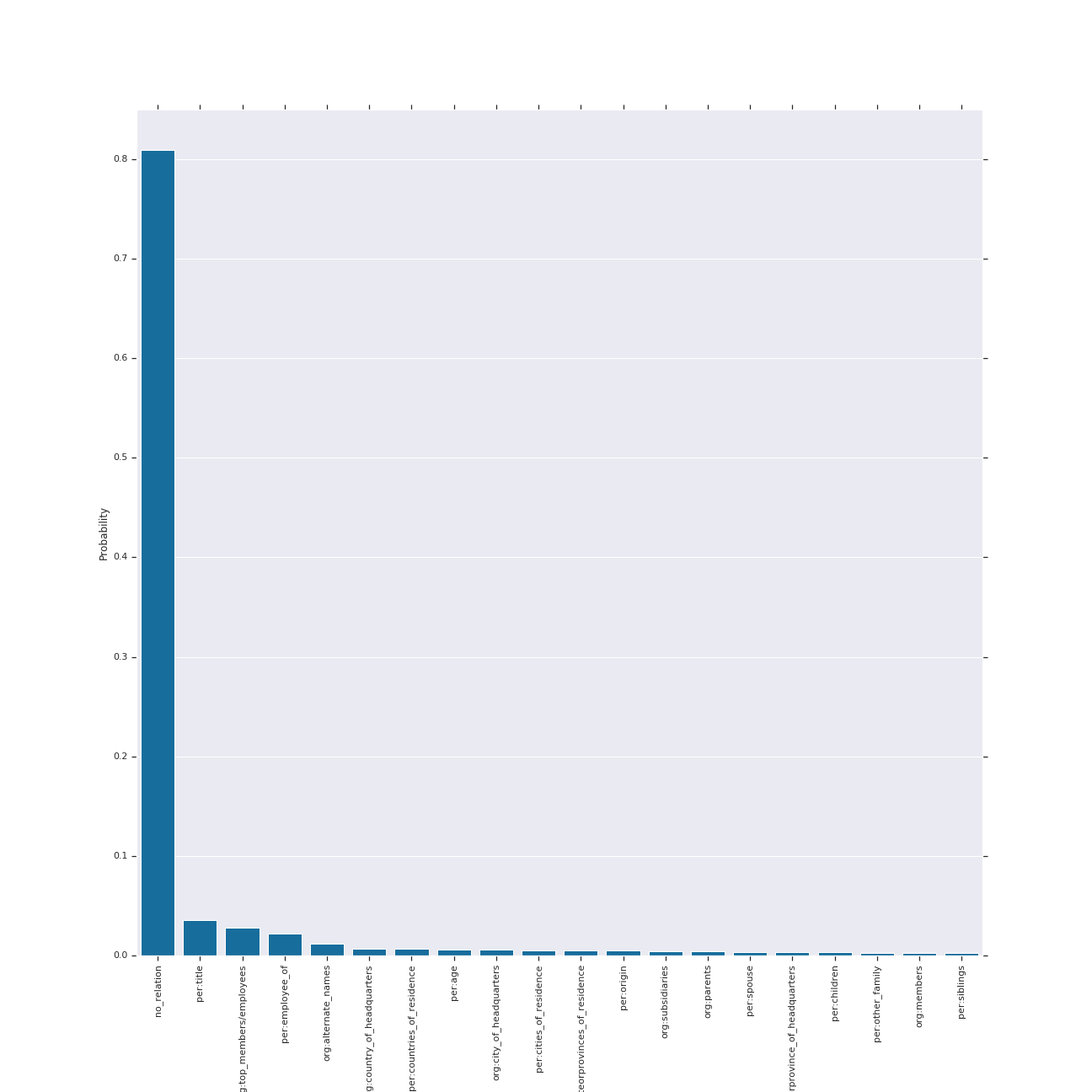}}%
  \subfloat[\webred{}]{\includegraphics[width=0.5\linewidth]{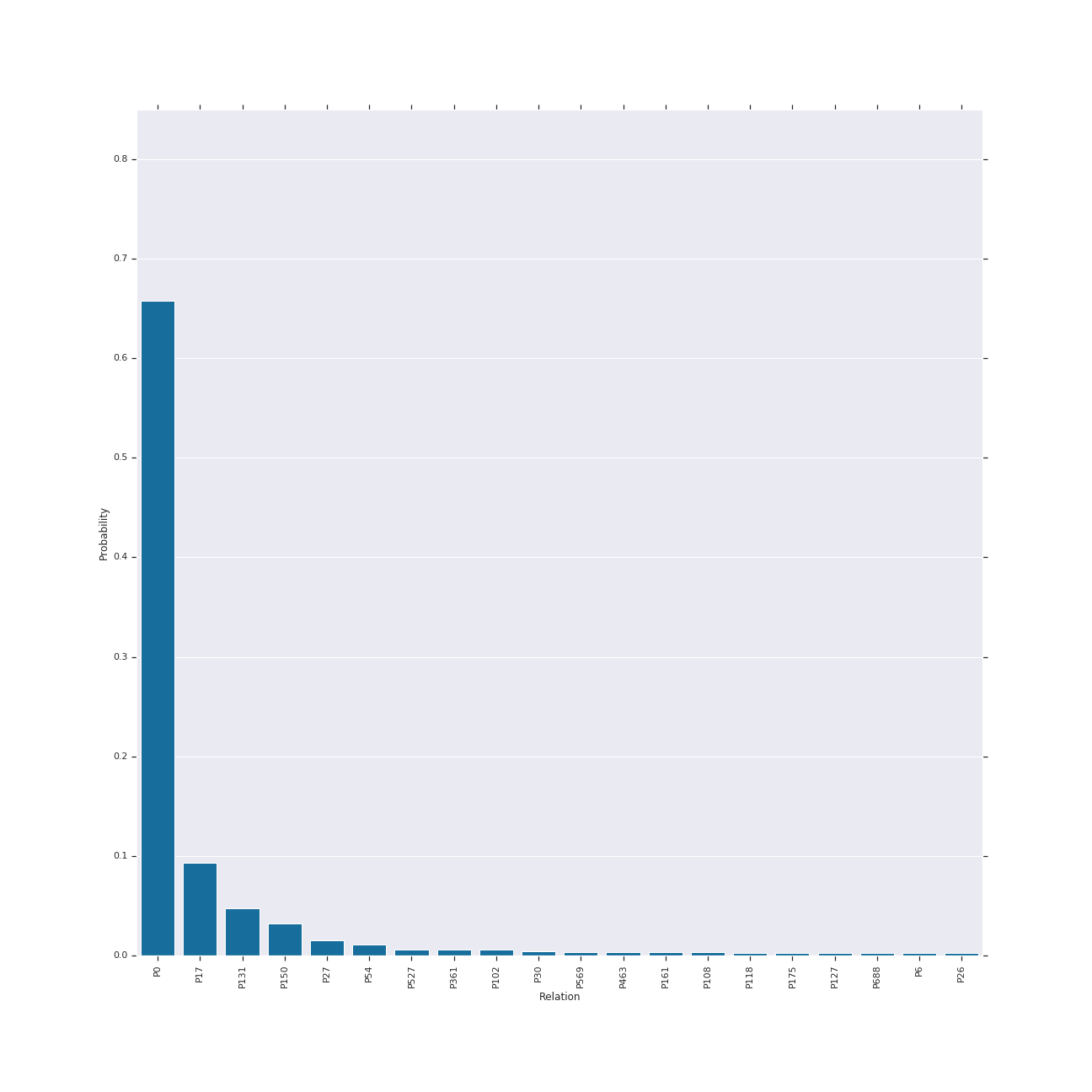}}%
\caption{The distribution of the examples per relation type (top-10 by frequency)}
\label{fig:appendix_rel_dist}
\end{figure*}

\begin{figure*}
  \centering
  \subfloat{\includegraphics[width=1.0\linewidth]{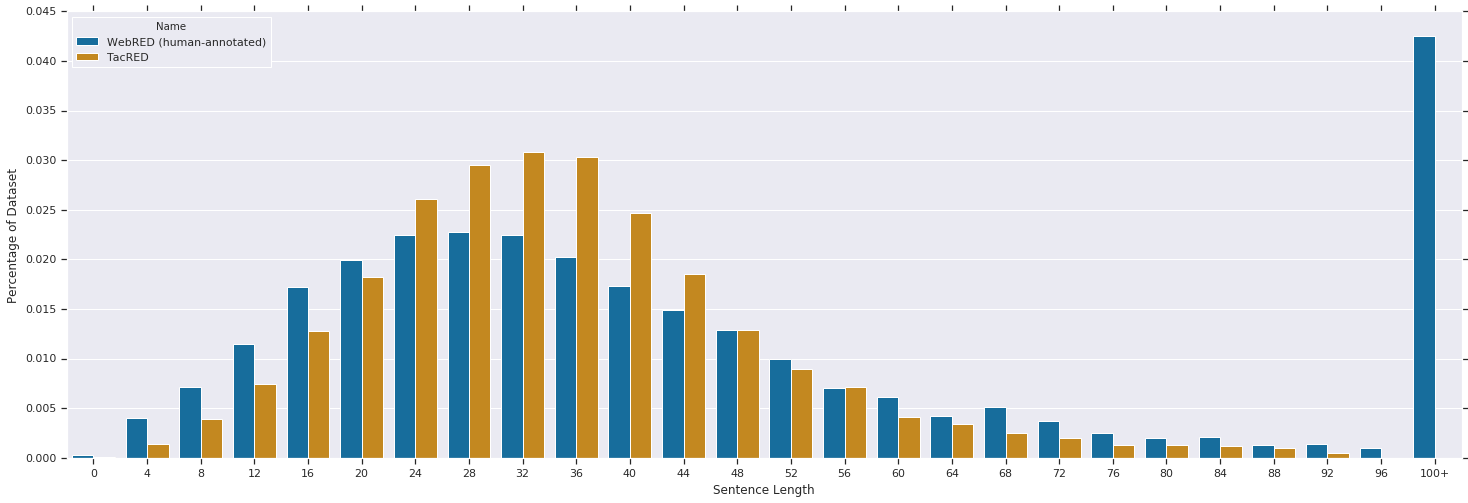}}
  \caption{Distribution of sentence lengths (in tokens) in \webred and TacRED.}
\label{fig:appendix_sentence_dist}
\end{figure*}

Figure \ref{fig:appendix_rel_dist} shows the frequency of the number of examples available per relation type (top-10 by frequency) for our dataset \webred, and TacRED. Although they follow a similar trend, \webred contains more positive examples for relations across all types, aside from having a higher number of examples in total. The relations in TacRED skew towards economic and political attributes of organizations or people, but the \webred distribution accommodates relations pertaining to biographical, geographical, and scientific topics. For example, there are no TacRED equivalents for concepts of authorship, performance, or relationships between countries. Similarly, Figure \ref{fig:appendix_sentence_dist} shows the distribution of sentence lengths found in \htwentyone and TacRED. \htwentyone dataset has an average sentence length of 41.9 tokens, whereas TacRED has an average sentence length of 36.4 tokens. \webred exposes models to sentences with a more varied distribution of lengths and a larger set of relations.

\subsection{Domains} \label{appendix:web_domains}
Table \ref{tab:web_domains} lists publicly-available domains that were crawled to form the text corpus of our dataset. They span a variety of text styles and topics including forms, wikis, scientific articles, and news of many types including politics, sports, science, nature, technology etc. A group of 10 people were surveyed and asked to select from a list of web-domains that they thought published articles with high linguistic quality and factual accuracy. 153 domains were shortlisted to form the source of all text found in our corpus. Afterwards, we sampled web-pages from these domains and that formed the text corpus for \webred{}.

\begin{table*}[t]
\centering
\begin{tabular}{|l|l|l|l|}
\hline
\multicolumn{4}{|l|}{\textbf{Domains}}\\
\hline
spn.com & sportingnews.com & nesn.com & skysports.com \\ \hline
thehill.com & salon.com & wnd.com & newsmax.com \\ \hline
motherjones.com & arstechnica.com & 9to5mac.com & fool.com \\ \hline
businessinsider.com & ft.com & ibtimes.com & w3.org \\ \hline
theguardian.com & yelp.com & tripadvisor.com & mit.edu \\ \hline
gnu.org & wiley.com & nature.com & economist.com \\ \hline
cbssports.com & washingtonpost.com & forbes.com & nytimes.com \\ \hline
cnn.com & usatoday.com & reuters.com & foxnews.com \\ \hline
cnbc.com & people.com & espn.com & cbsnews.com \\ \hline
bloomberg.com & newsweek.com & chicagotribune.com & seekingalpha.com \\ \hline
bleacherreport.com & vox.com & variety.com & nbcnews.com \\ \hline
eonline.com & latimes.com & theverge.com & marketwatch.com \\ \hline
nj.com & billboard.com & wsj.com & npr.org \\ \hline
si.com & hollywoodreporter.com & ajc.com & huffingtonpost.com \\ \hline
cnet.com & time.com & miamiherald.com & mercurynews.com \\ \hline
freep.com & usnews.com & nypost.com & ew.com \\ \hline
mashable.com & usmagazine.com & bostonglobe.com & startribune.com \\ \hline
tampabay.com & fortune.com & azcentral.com & politico.com \\ \hline
kansascity.com & cleveland.com & nbcsports.com & sfchronicle.com \\ \hline
mlive.com & techcrunch.com & chron.com & charlotteobserver.com \\ \hline
dallasnews.com & baltimoresun.com & theatlantic.com & qz.com \\ \hline
sacbee.com & today.com & oregonlive.com & orlandosentinel.com \\ \hline
suntimes.com & thedailybeast.com & nydailynews.com & boston.com \\ \hline
washingtontimes.com & denverpost.com & newyorker.com & nola.com \\ \hline
slate.com & wired.com & newsday.com & engadget.com \\ \hline
deadspin.com & gizmodo.com & zdnet.com & sltrib.com \\ \hline
fastcompany.com & syracuse.com & post-gazette.com & voanews.com \\ \hline
ocregister.com & venturebeat.com & sciencedaily.com & foxsports.com \\ \hline
livescience.com & bostonherald.com & pcmag.com & pcworld.com \\ \hline
inc.com & foxbusiness.com & barrons.com & sciencedirect.com \\ \hline
\end{tabular}
\caption{Domains used to form the text corpus of our dataset.}
    \label{tab:web_domains}
\end{table*}

\end{document}